%% file: acl_latex.tex
\pdfoutput=1

\documentclass[11pt]{article}

\usepackage[final]{acl}
\newif\ifarxiv
\arxivfalse 

\usepackage{times}
\usepackage{latexsym}

\usepackage[T1]{fontenc}

\usepackage[utf8]{inputenc}

\usepackage{microtype}

\usepackage{inconsolata}

\usepackage{graphicx}

\usepackage{booktabs}
\usepackage{paralist}
\usepackage{comment}

\title{EmoNews: A Spoken Dialogue System for Expressive News Conversations}

\author{
  Ryuki Matsuura$^1$\thanks{Equal contributions} ~~~
  Shikhar Bharadwaj$^{1*}$ ~~~
  Jiarui Liu$^{1*}$ ~~~
  Dhatchi Kunde Govindarajan$^{1*}$ \\
  ~Carnegie Mellon University \\
  ~\texttt{\{rmatsuur, sbharad2, jiaruil5, dkundego\}@andrew.cmu.edu}
}

\begin{document}
\maketitle

\input{sections/00_Abstract}
\input{sections/01_Introduction}

\input{sections/03_Method}

%
\input{sections/04_Experiment}

\input{sections/05_Conclusion}

\bibliography{custom}

\appendix



\end{document}

%% file: sections/00_Abstract.tex
\begin{abstract}
We develop a task-oriented spoken dialogue system (SDS) that regulates emotional speech based on contextual cues to enable more empathetic news conversations.
Despite advancements in emotional text-to-speech (TTS) techniques, task-oriented emotional SDSs remain underexplored due to the compartmentalized nature of SDS and emotional TTS research, as well as the lack of standardized evaluation metrics for social goals.
We address these challenges by developing an emotional SDS for news conversations that utilizes a large language model (LLM)-based sentiment analyzer to identify appropriate emotions and PromptTTS to synthesize context-appropriate emotional speech.
We also propose subjective evaluation scale for emotional SDSs and judge the emotion regulation performance of the proposed and baseline systems.
Experiments showed that our emotional SDS outperformed a baseline system in terms of the emotion regulation and engagement.
These results suggest the critical role of speech emotion for more engaging conversations.
All our source code is open-sourced.\footnote{\url{https://github.com/dhatchi711/espnet-emotional-news/tree/emo-sds/egs2/emo_news_sds/sds1}}
\end{abstract}

%% file: sections/01_Introduction.tex
\section{Introduction}
\label{sec:intro}

\begin{figure}[t]
    \centering
    \includegraphics[width=\columnwidth]{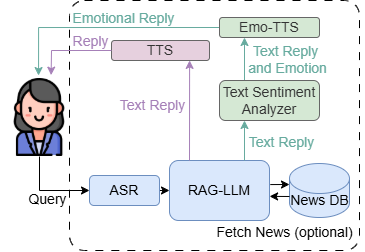}
    \caption{System architecture. Proposed system uses emoTTS and sentiment analyzer.}
    \label{fig:system_block_diag}
    \vspace{-5mm}
\end{figure}

In this work, we develop a task-oriented SDS that can regulate emotional TTS based on contextual cues (emotional SDS) to enable more empathetic news conversations. 
Task-oriented SDSs must balance task- and social-goals to create engaging interactions~\cite{clavel_socio-conversational_2022}, with emotional speech regulation being crucial among social-goals. For instance, synthesizing "sad" speech for tragic earthquake news can foster user empathy and engagement. Appropriately managing emotional tone can thus enhance both user perception and overall experience~\cite{kurata_development_2024, concannon_measuring_2024}. To support such needs, the field of affective computing has developed emotional TTS techniques, which generate emotionally expressive speech by adjusting acoustic features like cadence, intensity, and pitch. Recent emotional TTS systems have achieved high-quality oral emotional expressions~\cite{Cho2021MultiSpk, Wang2023FineTTS, bott_controlling_2024}.

However, despite these advances, task-oriented emotional SDSs remain underexplored. This is primarily because socio-conversational research has been compartmentalized~\cite{clavel_socio-conversational_2022}, with SDS and emotional TTS developing separately and lacking an integrated framework. Moreover, evaluating social-goals like emotional speech regulation is difficult~\cite{kurata_development_2024}, as these goals are multidimensional and lack clear definitions, leading to few established evaluation metrics. Thus, the gap between emotional TTS capabilities and their effective integration into SDS highlights an important area for further research.

We develop a task-oriented emotional SDS and propose its evaluation method. 
Specifically, we focus on news summarization and Q\&A as a target task due to its extensive prior studies. 
For emotional speech regulation, we adopt a cascade SDS architecture. 
We employ a PromptTTS~\cite{guo_prompttts_2022} fine-tuned on the \textit{ESD} dataset~\cite{zhou_emotional_2022} as our emotional TTS model. 
For evaluation, we use an empathy scale originally ~\cite{concannon_measuring_2024} and assess the SDS’s ability to regulate emotional speech. 
Additionally, we manually evaluate both the system’s emotional speech regulation and task achievement~\cite{walker_paradise_1997}, comparing it with SDSs that employ non-emotional TTS.

Through this study, we contribute to the studies on socio-conversational system by:
\begin{inparaenum}[(i)]
    \item providing a method for developing emotional SDS; and
    \item proposesing an evaluation method of emotional SDS.
\end{inparaenum}

%% file: sections/03_Method.tex
\section{System Design and Method}
\label{sec: problem formulation}



As depicted in Figure~\ref{fig:system_block_diag}, we develop an emotional SDS for news conversations using a cascade architecture, building on a strong baseline \cite{espnetSDS} by adding emotional awareness via sentiment-guided synthesis.
Below, we describe the core components of both systems.

The baseline system includes three modules: ASR, LLM, and TTS. The ASR transcribes user speech to text, which is encoded and compared against a News Database for relevant article retrieval. The LLM generates a response based on both the transcript and retrieved news snippets, and the TTS outputs spoken responses in a default tone. 
We utilize Retrieval Augmented Generation (RAG) in our system.
The core ASR and RAG-LLM module are shared: the ASR transcript is passed to a RAG language model that selectively retrieves news to ground its replies, using dynamic in-context prompting for adaptability.

Our proposed system enhances the baseline with a Sentiment Analyzer that infers the emotional tone (\texttt{neutral}, \texttt{happy}, \texttt{sad}, \texttt{angry}, or \texttt{surprised}) from the LLM’s text response. The emotion tag is fed to PromptTTS, an emotional TTS module that conditions speech synthesis on both text and emotion, producing expressive and empathetic responses. Compared to the emotionally neutral baseline, our system delivers more human-like, engaging interactions through sentiment understanding and emotional prosody.

%% file: sections/04_Experiment.tex
\begin{table*}[t]
\centering
\resizebox{\textwidth}{!}{
    
    \begin{tabular}{ll}
    \toprule
    Metric                                     & Item                                                                                                                                                   \\ \midrule
    RAG Evaluation                             & The system was helpful to understand the retrieved news.                                                                                               \\
    Task Achievement 1 (Usefulness)            & The news that the system retrieved matched the information you wanted to know.                                                                         \\
    Task Achievement 2  (Response Consistency) & The system consistently responded according to the retrieved news.                                                                                     \\
    Speech Emotion Appropriateness             & The system seemed to vary its emotional state of speech to demonstrate expressiveness \\
                                               & and modify its responses to accommodate the mood of the context. \\
    Engagement                                 & Did you have favorable feelings toward the one you were talking to?                                                                                    \\
                                               & Did you feel a sense of familiarity with the one you were talking to?                                                                                  \\
                                               & Did you feel that the system you were talking to understood the mood of contexts? \\\bottomrule       
    \end{tabular}
    }
    \caption{Emotional SDS Evaluation Questionnaire.}
    \label{tab:questionnaire}
\end{table*}

\begin{figure*}[!t]
    \centering
    \includegraphics[width=\textwidth,height=0.2\textheight]{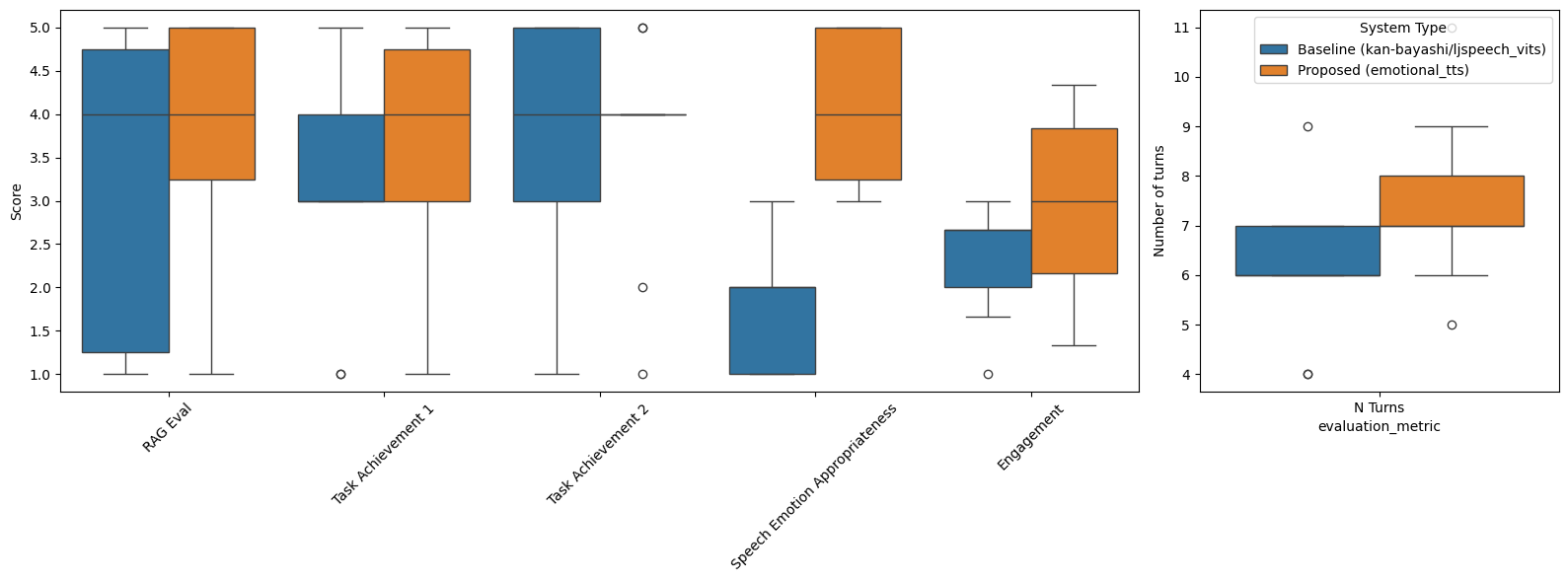}
    \caption{Comparison of Evaluation Metrics by System Type.}
    \label{fig:fig1}
    \vspace{-5mm}
\end{figure*}

\begin{table}[!t]
\centering
\resizebox{\columnwidth}{!}{
\begin{tabular}{lccc}
\toprule
\textbf{Metric} & \textbf{U} & \textbf{p-value} & \textbf{Cohen's d} \\
\midrule
RAG Evaluation        & 42.5  & 0.580       & 0.301 \\
Task Achievement 1    & 42.0  & 0.558       & 0.206 \\
Task achievement 2    & 51.0  & 0.968       & 0.070 \\
Speech Emotion Appropriateness         & 1.5   & $<0.001$    & 3.070 \\
Engagement            & 27.5  & 0.090       & 0.824 \\
N Turn               & 26.5  & 0.073       & 0.831 \\
\bottomrule
\end{tabular}
}
\caption{Statistical Comparison Between Baseline and Proposed Systems}
\label{tab:stats}
\vspace{-0.5cm}
\end{table}

\section{Experiments}

To evaluate the extent to which the proposed method can control proper speech emotion, we compared it with a baseline system using human subjective judgments.

\vspace{-.1cm}
\subsection{Datasets}

For emotional TTS fine-tuning, we used the English portion of the ESD dataset~\cite{zhou_emotional_2022}, splitting 17,500 utterances into training, validation, and evaluation subsets across five emotions.
For sentiment analyzer fine-tuning, we used GoodNewsEveryone~\cite{bostan_goodnewseveryone_2020} and GoEmotions~\cite{demszky_goemotions_2020}, mapping their emotion tags to five target categories following~\citet{koufakou_towards_2024}.
As the news database for retrieval, we used Free News\footnote{\href{FreeNews}{https://github.com/Webhose/free-news-datasets}}, filtering for English articles and embedding news titles with Chroma\footnote{\href{Chroma}{https://www.trychroma.com/}} and Sentence Transformers~\cite{reimers-gurevych-2019-sentence}.

\subsection{System Setups}

\paragraph{Proposed System} We used Whisper Large for ASR, LLaMA 3.2 1B for the language model, and a sentence transformer for retrieving the top 1 relevant news. For emotional TTS, we fine-tuned PromptTTS (pre-trained on LJSpeech). Our preliminary analysis showed that its quality was comparable to FastSpeech~\cite{ren_fastspeech_2019} and VITS~\cite{kim_conditional_2021} in terms of UTMOS, DNSMOS, PLCMOS, and WER, and qualitative analysis confirmed clear emotional variation. For the sentiment analyzer, we fine-tuned a distilled RoBERTa model (batch size 8, learning rate 0.00001, 4 epochs) after finding that prompt-based LLM approaches tended to over-predict sadness and surprise, achieving better performance than~\citet{koufakou_towards_2024}.

\paragraph{Baseline System} The baseline system shared the same modules as the proposed system, except the sentiment analyzer and a VITS model pre-trained on LJSpeech instead of emotional TTS.

\subsection{Metrics}

We create a seven-item questionnaire in Table~\ref{tab:questionnaire}, using a 5-point Likert scale (1 = strongly disagree, 5 = strongly agree). The first item assesses RAG performance on relevance and coherence, while the second and third address task achievement~\cite{walker_paradise_1997}: system helpfulness in understanding retrieved news and consistency of responses. The fourth item measures speech emotion appropriateness, adapted from empathy scales for dialogue systems~\cite{concannon_measuring_2024}. The last three items assess user engagement, based on Kurata et al.'s questionnaire~\cite{kurata_development_2024}. We also recorded the number of SDS turns as an additional engagement indicator~\cite{aoyama_conceptualization_2024}.

\vspace{-.2cm}
\subsection{Evaluation Procedure}
We collect 20 conversation samples by conducting 10 dialogues with each system. To avoid bias, emotion tags predicted by the sentiment analyzer were hidden from the SDS interface. We test differences in mean scores using Mann-Whitney U tests ($\alpha = .05$) due to the small sample size, and calculate Cohen’s d for effect sizes~\cite{cohen_statistical_2013}. 
We assess the internal consistency of the three engagement items using Cronbach’s alpha, which was .860, indicating substantial reliability; thus, we averaged them into a single engagement score.

\vspace{-0.25cm}
\subsection{Results and Discussion}

Figure~\ref{fig:fig1} shows the boxplots of human-judgment scores. The proposed system significantly outperformed the baseline in speech emotion appropriateness with a large effect size ($d=3.070$; 4.100 vs. 1.700), confirming its ability to control emotions according to context.
Although engagement scores and the number of turns showed no significant differences, both had large effect sizes (d = 0.824, 0.831), suggesting that emotional control may promote more engaging conversations~\cite{concannon_measuring_2024}. However, the mean engagement score remained moderate (around 3), possibly due to abrupt, discrete emotional shifts without considering prior conversational context.
Finally, no significant differences were observed in RAG performance or task achievement, and both systems scored around 3, indicating room for improvement in task-goal fulfillment.

%% file: sections/05_Conclusion.tex
\vspace{-.2cm}
\section{Conclusion}
\label{sec: conclusion}

We presented an emotional SDS that enhances empathetic interactions in task-oriented news conversations by combining a sentiment analyzer with PromptTTS for dynamic emotional speech generation. Our system integrates sentiment-driven emotional control within a prompt-based architecture, improving emotion appropriateness and engagement without compromising task performance. Its modular design enables easy adaptation to other domains, supporting broader development of emotionally aware conversational agents.